\documentclass[runningheads]{llncs}
\usepackage[T1]{fontenc}
\usepackage{graphicx}
\usepackage{amsmath,amssymb}
\usepackage{booktabs}
\usepackage{multirow}
\usepackage{algorithm}
\usepackage{algpseudocode}
\usepackage{url}
\usepackage{hyperref}
\usepackage[table]{xcolor}
\definecolor{lightblue}{RGB}{224,238,255}
\definecolor{lightgreen}{RGB}{224,245,228}

\begin{document}
\title{Dual Spatial-Temporal Attribution: Architecture-Aligned Post-Hoc Explainability for Recurrent Graph Anomaly Detection}
\titlerunning{Dual Spatial-Temporal Attribution for Dynamic Graph Anomaly Detection}
\author{Iyad Assaad Nekka\inst{1} \and
Hamida Seba\inst{2} \and
Khaled-Walid Hidouci\inst{1} \and
Karima Amrouche\inst{1}}
\authorrunning{I. A. Nekka et al.}
\institute{LCSI Laboratory, National Higher School of Computer Science (ESI), Algiers, Algeria\\
\email{\{i\_nekka, w\_hidouci, k\_amrouche\}@esi.dz}
\and
Universit\'e Claude Bernard Lyon 1, Lyon, France\\
\email{hamida.seba@univ-lyon1.fr}}
\maketitle
\begin{abstract}
Deep learning detectors for anomalies in dynamic graphs have reached
strong accuracy, yet they remain opaque: when an edge is flagged, the
analyst receives a score but no reason. This opacity is untenable in
the cooperative, regulated information systems where such detectors
are deployed, where automated decisions must be auditable and
trustworthy. We address this gap for AddGraph, the foundational
GCN+GRU framework for edge-level anomaly detection in dynamic graphs,
which to our knowledge has never been equipped with any form of
explainability. We present a strictly post-hoc explainability
framework, X-AddGraph, built on a Dual Spatial-Temporal Attribution
(DSTA) mechanism whose three components are each aligned with one of
AddGraph's architectural modules: a gradient-based relevance
attribution over the current adjacency structure (spatial), a direct
reading of the contextual attention weights already computed during
inference (short-term temporal, at zero additional cost), and a
gradient rollback through the recurrent hidden states (long-term
temporal). Because the detector is frozen, detection performance is
preserved exactly ($\Delta$AUC $= 0$, verified empirically to ten
decimal places). On the UCI Message benchmark, our trained AddGraph
baseline reaches an average per-snapshot AUC of 0.8705, exceeding the
originally published result; X-AddGraph reproduces every score
identically while adding explanations where none existed. Evaluated
across four edge populations---confident true positives,
low-confidence true positives, false positives, and random
samples---the long-term attribution identifies historical snapshots
carrying significantly more counterfactual signal than random
selection (0.127 vs.\ 0.074), a capability that no spatially-blind
explainer can provide. We release our implementation for full
reproducibility.

\end{abstract}
\section{Introduction}
\label{sec:intro}

Modern cooperative information systems---financial transaction
networks, enterprise communication platforms, distributed
infrastructures---generate continuous streams of interactions that
are most naturally modeled as \emph{dynamic graphs}: sequences of
graph snapshots in which nodes and edges appear, evolve, and vanish
over time. Detecting anomalous edges in these streams is a critical
capability: a fraudulent transaction, a lateral-movement connection
in an intrusion, or a coordinated manipulation campaign all manifest
as edges that deviate from the system's learned normality.

Deep learning has become the dominant paradigm for this task.
AddGraph~\cite{zheng2019addgraph}, presented at IJCAI 2019, was among
the first and remains among the most influential frameworks: it
combines a graph attention network for structural encoding with a
contextual attention-based module and a Gated Recurrent Unit (GRU)
for temporal integration, establishing the GCN+GRU paradigm that
subsequent methods such as StrGNN~\cite{cai2021structural} and
EvolveGCN~\cite{pareja2020evolvegcn} have followed.

Yet AddGraph, like the family it founded, is a black box. When it
flags an edge as anomalous, it produces a probability and nothing
else. An analyst cannot determine whether the alarm is driven by the
edge's structural neighborhood in the current snapshot, by a
suspicious pattern within the recent attention window, or by
long-term historical memory accumulated in the recurrent state. In
the cooperative and regulated environments that motivate this
conference---where decisions must be justified to auditors,
regulators, and affected parties---a detector that cannot explain
itself is difficult to trust, audit, or act upon. To our knowledge,
no prior work has equipped AddGraph, or any GCN+GRU dynamic graph
anomaly detector, with post-hoc explainability.

The challenge is architectural. AddGraph's anomaly score is the
product of three \emph{coupled} components: a graph convolution that
propagates the \emph{previous hidden state} (not raw features)
through the current adjacency; a contextual attention block (CAB)
that summarizes a sliding window of past states; and a GRU that fuses
both into the state used for scoring. A single generic attribution
method cannot disentangle these three signal paths. A complete
explanation must answer three distinct questions: \emph{which
neighbors}, \emph{which recent step}, and \emph{which historical
snapshot} drove the decision.

\subsubsection{Contributions.} This paper makes the following
contributions:
\begin{enumerate}
\item We present \textbf{X-AddGraph}, to our knowledge the first
  post-hoc explainability framework for AddGraph and for the GCN+GRU
  paradigm of dynamic graph anomaly detection.
\item We introduce \textbf{Dual Spatial-Temporal Attribution (DSTA)},
  a three-component mechanism in which each component is aligned with
  one architectural module of the detector: gradient-based relevance
  over the adjacency (spatial), direct reading of the CAB attention
  distribution (short-term, at zero additional cost), and gradient
  rollback through the GRU hidden states (long-term).
\item We show that the framework is \textbf{strictly post-hoc}:
  detection AUC is preserved exactly by construction, and we verify
  $\Delta\mathrm{AUC} = 0$ empirically to ten decimal places.
\item We evaluate on a broadened protocol covering \textbf{four edge
  populations} (confident true positives, low-confidence true
  positives, false positives, and random samples) across multiple
  seeds, comparing against a flat-gradient baseline, and demonstrate
  that the long-term attribution identifies historical snapshots
  carrying substantially more counterfactual signal than random
  selection---a capability structurally unavailable to
  spatially-blind explainers.
\item We provide a qualitative walkthrough of real flagged anomalies
  and release our implementation publicly for full
  reproducibility.\footnote{\url{https://github.com/iyadnekka/x-addgraph}}
\end{enumerate}

\subsubsection{Paper organization.} Section~\ref{sec:background}
reviews AddGraph's architecture and formulates the explainability
problem. Section~\ref{sec:related} positions our work within the
explainability literature. Section~\ref{sec:method} presents the
DSTA mechanism and the X-AddGraph algorithm.
Section~\ref{sec:eval} describes the evaluation protocol, and
Section~\ref{sec:results} reports quantitative and qualitative
results. Section~\ref{sec:discussion} discusses implications for
trustworthy information systems, generalization, and limitations.
Section~\ref{sec:conclusion} concludes.

\section{Background: The AddGraph Architecture}
\label{sec:background}

\subsection{Problem Formulation}
A dynamic graph is a sequence of snapshots
$\mathcal{G} = \{G^t\}_{t=1}^{T}$ with $G^t = (V, E^t)$, where each
edge $e = (i, j, w) \in E^t$ connects nodes $i$ and $j$ with weight
$w$ at time $t$. The detector computes an anomaly score
$f(e) \in [0,1]$ for every edge; edges whose score exceeds a
threshold $\tau$ are flagged. The explainability problem we address
is: \emph{given a flagged edge, produce a faithful, human-readable
account of which structural and temporal factors produced its score,
without modifying the detector.}

\subsection{Three Coupled Components}
AddGraph maintains a hidden state matrix
$H^t \in \mathbb{R}^{n \times d}$ across snapshots. At each time $t$,
three operations update it.

\paragraph{Structural encoding.} A graph attention network takes the
\emph{previous} hidden state $H^{t-1}$ and the current adjacency
$A^t$ and produces the structural summary
$\mathit{Current}^t = \mathrm{GAT}(H^{t-1}, A^t)$. Crucially, the
input is accumulated temporal memory, not raw features: the spatial
and temporal dimensions are coupled from the first step.

\paragraph{Short-term attention (CAB).} Over a sliding window of size
$\omega$, for each node $i$ the contextual attention block computes
\begin{equation}
a^{t}_{i} = \mathrm{softmax}\!\left(r^{\top}
\tanh\!\big(Q\,[h_i^{t-\omega};\ldots;h_i^{t-1}]^{\top}\big)\right)
\in \mathbb{R}^{\omega},
\label{eq:cab}
\end{equation}
and the short-term summary
$\mathit{Short}^t_i = \sum_{s=1}^{\omega} a^t_i[s]\,h_i^{t-\omega+s-1}$.
The attention vector $a^t_i$ is computed during the standard forward
pass and constitutes a normalized, model-endogenous importance
distribution over the window---a fact our method exploits directly.

\paragraph{Long-term integration (GRU).} A GRU fuses
$\mathit{Current}^t$ and $\mathit{Short}^t$ through update and reset
gates into the new state $H^t$. History prior to the window is
encoded implicitly in the recurrence.

\paragraph{Scoring.} For an edge $(i, j, w)$,
\begin{equation}
f(i,j,w) = w \cdot \sigma\!\left(\beta
\left\| a \odot h_i^t + b \odot h_j^t \right\|_2^2 - \mu\right),
\label{eq:score}
\end{equation}
where $a, b, \beta, \mu$ are learned or fixed parameters and
$\sigma$ denotes the logistic function. Note that $\sigma$ is applied
exactly once, inside Eq.~\eqref{eq:score}; all attribution procedures
in this paper operate on this final squashed score.

\section{Related Work}
\label{sec:related}

\paragraph{Anomaly detection in dynamic graphs.}
AddGraph~\cite{zheng2019addgraph} established the GCN+GRU template.
StrGNN~\cite{cai2021structural} applies it at the enclosing-subgraph
level; EvolveGCN~\cite{pareja2020evolvegcn} evolves the convolution
weights themselves through a recurrent cell;
TADDY~\cite{liu2021taddy} replaces the pipeline with a Transformer
over sampled neighborhood tokens. Surveys confirm deep methods as the
dominant paradigm for this task~\cite{ekle2024survey}.

\paragraph{Post-hoc explainability for GNNs.}
GNNExplainer~\cite{ying2019gnnexplainer} learns soft masks over edges
and features for static GNNs; PGExplainer~\cite{luo2020pgexplainer}
parameterizes the mask generator globally;
GraphSVX~\cite{duval2021graphsvx} extends Shapley values to graph
inputs; GRAM~\cite{pope2019explainability} computes gradient-weighted
attention maps. All assume a static model and provide no mechanism
for temporal attribution: applied to AddGraph, each would explain at
most the structural component while silently ignoring the CAB and GRU
contributions.

\paragraph{Explainability for dynamic and temporal GNNs.}
Closest to our setting, DGExplainer~\cite{xie2022dgexplainer} derives
layer-wise relevance propagation rules for GCN$\to$GRU pipelines and
empirically outperforms GNNExplainer, PGExplainer, SubgraphX,
T-GNNExplainer~\cite{xia2023tgnnexplainer}, and
DyExplainer~\cite{wang2023dyexplainer}. However, DGExplainer's
propagation rules assume a direct GCN$\to$GRU composition with no
intermediate attention: AddGraph's CAB module breaks this assumption,
and DGExplainer provides no rule for a softmax attention layer
interposed between convolution and recurrence.
DyExplainer~\cite{wang2023dyexplainer} takes the intrinsic route,
training a self-explainable backbone with sparse attentions---an
approach that requires retraining and therefore cannot preserve a
deployed detector's behavior. Recent work has also begun addressing
heterogeneous settings: Han et
al.~\cite{han2026relation} explain anomalies in dynamic
heterogeneous graphs via relation evolution, underscoring the
community's growing consensus that detection without explanation is
operationally insufficient. Our work fills the specific gap left
open by this literature: an attribution framework whose components
are aligned, one-to-one, with the coupled GAT--CAB--GRU architecture,
requiring no retraining and no architectural modification.

\section{X-AddGraph: Dual Spatial-Temporal Attribution}
\label{sec:method}

\subsection{Design Principles}
Three principles follow directly from the architecture analysis.

\emph{(P1) Strictly post-hoc.} The framework operates on a frozen,
trained AddGraph. No retraining, no weight modification, no change to
the inference pipeline. Detection behavior is therefore preserved
\emph{by construction}.

\emph{(P2) Architecture-aligned decomposition.} A complete
explanation must answer three orthogonal questions, one per
architectural module. A method answering only one provides an
incomplete and potentially misleading account.

\emph{(P3) Exploit free signals.} The CAB attention weights
(Eq.~\ref{eq:cab}) are computed during every forward pass and already
form a normalized distribution over window steps. Reading them
directly is both computationally free and exactly faithful to the
model's own internal weighting---strictly preferable to
approximating the same quantity through an additional backward pass
through the softmax.

\subsection{Component 1: Spatial Attribution}
For a flagged edge $(i^*, j^*)$ at time $t^*$, we attribute the score
to entries of the adjacency by the input-weighted gradient
\begin{equation}
\phi^{\mathrm{sp}}_{uv} \;=\;
\Big|\, A^{t^*}_{uv}\cdot
\frac{\partial f(i^*,j^*,w^*)}{\partial A^{t^*}_{uv}} \Big|,
\qquad
\phi^{\mathrm{sp}}_{u} \;=\; \sum_{v}\phi^{\mathrm{sp}}_{uv}
+ \sum_{v}\phi^{\mathrm{sp}}_{vu}.
\label{eq:spatial}
\end{equation}
This gradient$\times$input form is the practical instantiation of
relevance propagation for the ELU-activated convolutional layer,
consistent with the rules derived in
DGExplainer~\cite{xie2022dgexplainer}, and correctly traverses the
coupled path: the gradient flows backward through the score function,
the GRU gates, and the graph attention layer in a single backward
pass. The top-$k$ nodes by $\phi^{\mathrm{sp}}_u$ form the structural
explanation $\mathcal{N}^*$.

\subsection{Component 2: Short-Term Attribution (Zero-Cost)}
The short-term attribution for edge $(i^*, j^*)$ is the endpoint
average of the CAB attention vectors,
\begin{equation}
\phi^{\mathrm{sh}}_s = \tfrac{1}{2}\big(a^{t^*}_{i^*}[s] +
a^{t^*}_{j^*}[s]\big), \qquad s = 1,\ldots,\omega,
\label{eq:short}
\end{equation}
with the most suspicious window step
$s^* = \arg\max_s \phi^{\mathrm{sh}}_s$. Since the softmax in
Eq.~\eqref{eq:cab} guarantees $\sum_s a^{t^*}_{i}[s] = 1$, the
attribution is a valid probability distribution requiring no
renormalization and \emph{no additional computation whatsoever}.

\subsection{Component 3: Long-Term Attribution}
History beyond the window is encoded in the recurrence. We attribute
the score to each pre-window snapshot by the Frobenius norm of the
gradient obtained through backpropagation through time,
\begin{equation}
g^{k} = \Big\| \frac{\partial f(i^*, j^*, w^*)}
{\partial H^{\,t^*-\omega-k}} \Big\|_F,
\qquad
\phi^{\mathrm{lo}}_{k} = \frac{g^{k}}{\sum_{k'} g^{k'}},
\qquad k = 1, \ldots, K,
\label{eq:long}
\end{equation}
where $K$ is the lookback horizon. The rollback naturally respects
the GRU's gating: snapshots whose influence was suppressed by the
reset gate receive proportionally small gradient signal. The most
influential historical snapshot is
$k^* = \arg\max_k \phi^{\mathrm{lo}}_k$.

\subsection{The DSTA Explanation Triplet}
Each flagged edge receives the triplet
\begin{equation}
\mathcal{E}(i^*, j^*) = \big(\mathcal{N}^*,\;
t^*{-}\omega{+}s^*,\; t^*{-}\omega{-}k^*\big),
\end{equation}
read in natural language as: \emph{``the anomaly is driven primarily
by connections to $\mathcal{N}^*$; the most suspicious recent
behavior occurred at window step $s^*$; the historical context most
responsible originates $k^*$ snapshots before the window.''}
Algorithm~\ref{alg:xaddgraph} summarizes the procedure.

\begin{algorithm}[t]
\caption{X-AddGraph: DSTA explanation generation}
\label{alg:xaddgraph}
\begin{algorithmic}[1]
\Require frozen AddGraph; flagged edges $\mathcal{F}$ at time $t^*$;
window $\omega$; top-$k$; lookback $K$
\Ensure DSTA triplet $\mathcal{E}(e)$ for each $e \in \mathcal{F}$
\For{each $e^* = (i^*, j^*, w^*) \in \mathcal{F}$}
  \State \textbf{Spatial:} one backward pass of $f$ w.r.t.\
         $A^{t^*}$; compute $\phi^{\mathrm{sp}}$
         via Eq.~\eqref{eq:spatial};
         $\mathcal{N}^* \gets$ top-$k$ nodes
  \State \textbf{Short-term:} read cached attention
         $a^{t^*}_{i^*}, a^{t^*}_{j^*}$; compute
         $\phi^{\mathrm{sh}}$ via Eq.~\eqref{eq:short};
         $s^* \gets \arg\max_s \phi^{\mathrm{sh}}_s$
         \Comment{zero cost}
  \State \textbf{Long-term:} for $k = 1..K$: one BPTT backward pass
         per lag; compute $\phi^{\mathrm{lo}}$ via
         Eq.~\eqref{eq:long}; $k^* \gets \arg\max_k
         \phi^{\mathrm{lo}}_k$
  \State $\mathcal{E}(e^*) \gets (\mathcal{N}^*,\,
         t^*{-}\omega{+}s^*,\, t^*{-}\omega{-}k^*)$
\EndFor
\State \Return $\{\mathcal{E}(e^*)\}$
\end{algorithmic}
\end{algorithm}

\subsection{Computational Cost}
Per flagged edge, the framework requires one forward--backward pass
for the spatial component, zero computation for the short-term
component, and $K$ backward passes for the long-term rollback. The
overhead applies only to the small fraction of edges exceeding the
detection threshold, and detection throughput itself is entirely
unaffected.

\section{Evaluation Protocol}
\label{sec:eval}

\subsection{Dataset and Detector Training}
We evaluate on \textbf{UCI Message}~\cite{opsahl2009clustering}, a
standard benchmark from the original AddGraph study: 1{,}899 nodes
and 59{,}835 timestamped edges from an online student community. We
follow the original protocol exactly: the first 50\% of the edge
stream forms the training graph; anomalous edges are injected into
the remaining stream at 5\%; snapshots contain 5{,}300 edges. We
train the faithful reference implementation for 35 epochs with the
original hyperparameters (hidden dimension 100, window $\omega = 2$,
margin $\gamma = 0.6$, four attention heads, selective negative
sampling with hard-negative resampling). Training converges from a
margin-loss plateau of 0.600 to 0.408. Our trained detector reaches
an \textbf{average per-snapshot AUC of 0.8705} (per-snapshot values
0.895/0.867/0.852/0.873/0.849/0.888), exceeding the originally
published 0.8083~\cite{zheng2019addgraph} and thereby providing a
strong, non-trivial detector to explain.

\subsection{Broadened Edge Populations}
Explainability evaluations that consider only high-confidence true
positives risk overstating performance on the easiest cases. We
therefore evaluate across \textbf{four edge populations}, ten edges
each: (i) \emph{confident true positives} (highest-scored injected
anomalies), (ii) \emph{low-confidence true positives} (correctly
flagged but near-threshold), (iii) \emph{false positives} (normal
edges the detector scored highest), and (iv) a \emph{uniformly random
sample}. All experiments are repeated over two random seeds, and we
report mean $\pm$ standard deviation. Each edge is explained using
the exact hidden state, adjacency, and history that produced its
detection score.

\subsection{Metrics and Baseline}
We report \textbf{Fidelity$^{+}$} (does the top-$k$ structural
explanation alone reproduce the score?), \textbf{Sparsity} (how
concentrated is the explanation relative to the full neighborhood?),
and per-explanation \textbf{runtime}. As a comparison method we
implement a \emph{flat-gradient baseline}: the same input-weighted
adjacency gradient, but with no temporal decomposition of any
kind---representative of what any static, spatially-oriented
explainer can offer when applied to this architecture.

To isolate the contribution of the temporal components---which a
spatially-blind method cannot produce at all---we introduce a
\textbf{temporal fidelity} test: for each explained edge, we compare
the counterfactual effect of the \emph{identified} window step /
historical snapshot against that of a \emph{randomly selected} one.
Random selection is the honest stand-in for a method with no temporal
mechanism. For the short-term test we replace the identified step
with the mean of the remaining window states and measure the score
change; for the long-term test we measure the divergence between the
counterfactual score at the identified lag and at a random lag.

\section{Results}
\label{sec:results}

\subsection{Detection Preservation: The Post-Hoc Guarantee}
Table~\ref{tab:detection} reports the central result. Because
X-AddGraph never touches the detector, its detection performance is
identical to AddGraph's by construction; we verify this empirically
by reproducing every flagged edge's score through the explanation
pipeline and measuring the maximum absolute deviation:
$\Delta = 0.0000000000$. Where AddGraph offers no explanation
capability of any kind (N/A), X-AddGraph provides the full DSTA
triplet at zero detection cost. The AUC of the explainer-side run
(0.8491) differs slightly from the training-side evaluation (0.8705)
only because the stochastic anomaly-injection procedure is re-seeded
when the test stream is rebuilt; the underlying detector and all its
scores are bit-identical, as the $\Delta$ verification confirms.

\begin{table}[t]
\centering
\caption{Detection and explainability: AddGraph vs.\ X-AddGraph on
UCI Message. X-AddGraph is strictly post-hoc; detection is preserved
exactly. N/A indicates the capability does not exist in the base
detector.}
\label{tab:detection}
\begin{tabular}{lcc}
\toprule
\textbf{Property} & \textbf{AddGraph} & \cellcolor{lightgreen}\textbf{X-AddGraph} \\
\midrule
\rowcolor{lightblue}
Avg.\ per-snapshot AUC & 0.8705 & \textbf{0.8705} \\
$\Delta$AUC (verified) & --- & \textbf{0.0000000000} \\
\rowcolor{lightblue}
Structural explanation & N/A & \checkmark \\
Short-term temporal explanation & N/A & \checkmark \\
\rowcolor{lightblue}
Long-term temporal explanation & N/A & \checkmark \\
Fidelity$^{+}$ (confident TPs) & N/A & \textbf{1.00 $\pm$ 0.00} \\
\rowcolor{lightblue}
Long-term attribution vs.\ random & N/A & \textbf{0.127 vs.\ 0.074} \\
\bottomrule
\end{tabular}
\end{table}

\subsection{Structural Fidelity and Sparsity Across Populations}
Table~\ref{tab:buckets} reports Fidelity$^{+}$ and Sparsity across
the four edge populations. The top-$k$ structural explanation
reproduces the anomaly score essentially perfectly on all true
positives (Fidelity$^{+} = 1.000 \pm 0.000$ to $1.001 \pm 0.004$) and
near-perfectly on false positives ($0.997 \pm 0.006$). Sparsity rises
monotonically from confidently flagged anomalies (0.000) to random
edges (0.750): edges that the detector confidently flags tend to
involve structurally isolated endpoints whose entire local
neighborhood is explanation-relevant, whereas explanations of
ordinary edges concentrate on a small fraction of a larger
neighborhood---itself an interpretable and operationally useful
signal. One structural property of the base detector deserves
explicit mention: because AddGraph's convolution operates on one
sparse snapshot at a time (mean within-snapshot degree 1.75), a
top-5 explanation frequently covers the full local neighborhood,
which is why structural fidelity saturates for both methods in
Table~\ref{tab:buckets}. Fidelity values marginally above 1 on the
random population indicate cases where the removed edges were
actively suppressing the anomaly signal.

\begin{table}[t]
\centering
\caption{Structural fidelity and sparsity across four edge
populations (mean $\pm$ std over 2 seeds, 10 edges per population).}
\label{tab:buckets}
\begin{tabular}{lccc}
\toprule
\textbf{Population} & \textbf{Fidelity$^{+}$ (ours)} &
\textbf{Fidelity$^{+}$ (flat grad.)} & \textbf{Sparsity} \\
\midrule
\rowcolor{lightblue}
Confident TP    & \textbf{1.001 $\pm$ 0.004} & 1.001 $\pm$ 0.004 & 0.000 \\
Low-conf.\ TP   & \textbf{1.000 $\pm$ 0.000} & 1.000 $\pm$ 0.000 & 0.133 \\
\rowcolor{lightblue}
False positives & \textbf{0.997 $\pm$ 0.006} & 0.997 $\pm$ 0.006 & 0.133 \\
Random          & \textbf{1.092 $\pm$ 0.280} & 1.092 $\pm$ 0.280 & 0.750 \\
\bottomrule
\end{tabular}
\end{table}

\subsection{Temporal Fidelity: What Spatially-Blind Methods Cannot Do}
Table~\ref{tab:temporal} reports the decisive comparison. The
historical snapshot identified by X-AddGraph's gradient rollback
carries a counterfactual divergence of \textbf{0.127}, against
\textbf{0.074} for a random pick---a 73\% relative advantage,
concentrated most strongly on confident true positives (0.332 vs.\
0.125). This is precisely the capability that no static explainer
possesses: a flat-gradient method has no mechanism for ranking
historical snapshots at all and is reduced to random selection on
this dimension. The long-term attribution is therefore not an
incremental improvement over an existing baseline but a categorical
addition to what can be explained.

\begin{table}[t]
\centering
\caption{Long-term temporal fidelity: counterfactual effect of the
identified historical snapshot vs.\ a random selection (mean over
seeds). Random selection represents the ceiling of any method lacking
a temporal attribution mechanism.}
\label{tab:temporal}
\begin{tabular}{lcc}
\toprule
\textbf{Long-term attribution} & \cellcolor{lightgreen}\textbf{Identified (ours)} & \textbf{Random} \\
\midrule
\rowcolor{lightblue}
All populations (mean)   & \textbf{0.127} & 0.074 \\
Confident TPs            & \textbf{0.332} & 0.125 \\
\bottomrule
\end{tabular}
\end{table}

\subsection{Runtime}
A full DSTA explanation takes 10.1\,s per flagged edge on a single
T4 GPU (spatial pass, cached attention read, and $K{=}5$ BPTT
rollbacks), against 1.65\,s for the flat-gradient baseline that
produces only the spatial component. The overhead applies
exclusively to flagged edges---a small fraction of the
stream---and leaves detection throughput untouched.

\subsection{Qualitative Walkthrough}
Explainability metrics alone do not convey operational value; we
therefore walk through three genuinely flagged anomalies from the
confident-TP population.

\emph{Example 1.} Edge (17, 1199), score 0.9911. The spatial
component attributes the alarm to the endpoint pair itself
\{1199, 17\}, indicating an isolated, structurally unprecedented
connection. The long-term component points to lag 4 (weight 0.475):
the historical context four snapshots before the window contributed
most to the memory state that rendered this edge anomalous.

\emph{Example 2.} Edge (490, 888), score 0.9544. Spatial drivers
\{888, 490, 34\}: a third node participates in the structural
context. The long-term component identifies the immediately
pre-window snapshot (lag 1, weight 0.473).

\emph{Example 3.} Edge (931, 698), score 0.6986. A lower-confidence
alarm with a richer structural explanation \{1263, 698, 1282, 244\}
and the strongest historical concentration of the three (lag 1,
weight 0.668)---the analyst learns that this alarm rests
predominantly on recent historical memory rather than on the
instantaneous topology.

In each case the analyst receives, alongside the score, a ranked
neighbor list and a concrete temporal locus---exactly the
information required to triage, audit, or dismiss the alarm.

\section{Discussion}
\label{sec:discussion}

\subsection{Implications for Trustworthy Cooperative Systems}
Anomaly detectors are increasingly embedded in cooperative
information systems where their outputs trigger consequential
actions: freezing transactions, revoking credentials, escalating
incidents. Regulatory frameworks increasingly require that such
automated decisions be explainable to auditors and affected parties.
X-AddGraph demonstrates that this requirement need not force a
trade-off against detection quality: because the framework is
strictly post-hoc, organizations can retrofit explainability onto
already-deployed, already-validated detectors with zero behavioral
risk---the deployed model's every decision remains bit-identical.
This deployment property is, in our view, as important as the
attribution quality itself for adoption in production information
systems.

\subsection{Generalization Beyond AddGraph}
DSTA's design principle---align each attribution component with one
architectural module---extends by construction to the GCN+GRU family
AddGraph founded. StrGNN~\cite{cai2021structural} shares the same
backbone at the subgraph level and is directly compatible with the
spatial and long-term components;
EvolveGCN~\cite{pareja2020evolvegcn} admits the same gradient
rollback through its recurrent weight evolution. The short-term
component transfers to any architecture exposing an internal
attention distribution, which is an increasingly common design
pattern.

\subsection{Limitations}
We note our limitations transparently. First, the discriminative
power of the short-term component is inherently bounded by
AddGraph's original window size $\omega = 2$, under which the
attention distribution has a single degree of freedom and remains
close to uniform in the trained model; a systematic study across
larger window sizes is planned for the final version of this work. Second, the present
evaluation covers one benchmark; extending to Digg and to
heterogeneous settings is planned, and we deliberately prioritized
depth of protocol (four edge populations, seed variance,
counterfactual temporal tests) over dataset breadth. Third, our
fidelity metrics are counterfactual proxies; ground-truth causal
evaluation would require benchmarks with annotated culprit
structures, which do not yet exist for dynamic graph anomaly
detection and which we view as an important direction for the
community.

\section{Conclusion}
\label{sec:conclusion}

We presented X-AddGraph, to our knowledge the first post-hoc
explainability framework for AddGraph and for the GCN+GRU paradigm of
dynamic graph anomaly detection. Its Dual Spatial-Temporal
Attribution mechanism aligns three attribution components with the
detector's three architectural modules, exploits the model's own
attention weights as a zero-cost short-term signal, and preserves
detection performance exactly---$\Delta\mathrm{AUC} = 0$, verified
to ten decimal places on a detector whose 0.8705 average AUC exceeds
the originally published result. Across four edge populations, the
long-term attribution identifies historical snapshots carrying 73\%
more counterfactual signal than random selection, a capability that
is categorically unavailable to spatially-blind explainers. Detection
without explanation is useful; detection with a faithful,
architecture-aligned explanation is auditable, trustworthy, and
actionable---the standard that cooperative information systems
increasingly demand.

\subsubsection{Acknowledgements.}
In accordance with the conference's policy on generative AI, we
disclose that the AI assistant Claude (Anthropic) was used to assist
with language editing of the manuscript and engineering of
experimental scaffolding code; all scientific ideas, the methodology,
the experiments, and the validation of all results are the authors'
own, who take full responsibility for the entire content.

\end{document}